\documentclass{article}
\usepackage{spconf,amsmath,graphicx}
\usepackage{subcaption}
\usepackage{amsfonts,epsfig,array,multirow,ltablex,tabularx,setspace,amssymb,multirow}
\usepackage{schemabloc,tikz}
\usepackage{cite}
\usepackage[justification=centering]{caption}
\usepackage[font={small}]{caption}
\usepackage{textcomp}
\usepackage{pifont}
\usepackage{bm}

\title{Design of Image-Matched Non-Separable Wavelet using Convolutional Neural Network}
\name{Naushad Ansari$^\ast$\thanks{$^\ast$Thanks to CSIR, Govt. of India for funding.}, Anubha Gupta, and Rahul Duggal}
\address{SBILab, Deptt. of ECE, IIIT-Delhi, India}
\begin{document}
%
\maketitle
\begin{abstract}
Image-matched nonseparable wavelets can find potential use in many applications including image classification, segmentation, compressive sensing, etc. This paper proposes a novel design methodology that utilizes convolutional neural network (CNN) to design two-channel non-separable wavelet matched to a given image. The design is proposed on quincunx lattice. The loss function of the convolutional neural network is setup with total squared error between the given input image to CNN and the reconstructed image at the output of CNN, leading to perfect reconstruction at the end of training. Simulation results have been shown on some standard images. 
\end{abstract}
\begin{keywords}
Nonseparable wavelet, Quincunx lattice, Convolutional Neural Network
\end{keywords}
\section{Introduction}
\label{Section for Into}
Over the past three decades, wavelets have been found utility in a number of applications spanning different disciplines including geoscience engineering, image processing applications, signal processing, etc. \cite{abry2015multiscale,pradhan2016data,tang2015hyperspectral,sakkari2015architecture,cirujeda20153d,
jiang2015isotropic,unser2014wavelets,puspoki2016design,kumar2007text}. In particular, choosing an appropriate wavelet in image segmentation, classification, compressive sensing, and similar applications, is generally a great challenge. There exist more number of standard wavelets for one-dimensional signals than the nonseparable wavelets for images because the latter are difficult to design. Hence, often one-dimensional wavelets are used as separable wavelets in image processing applications. However, there may be inter-dependence of information along different directions. Hence, a non-separable image-matched wavelet may provide better results in applications compared to a standard separable wavelet.

Although a number of methods have been proposed for the design of multidimensional wavelets \cite{zhou2005multidimensional,lu2007multidimensional,ruedin2002construction,wei2000new,puspoki2016design,ward2015optimal,
kovacevic1995nonseparable,kovacevic1992nonseparable,tay1993flexible}, these have, largely, been designed irrespective of signal of interest. In \cite{gupta2011two}, statistically matched non-separable wavelets have been designed for a class of images belonging to fractional Brownian field. The method first designs highpass filter using the signal statistics followed by the conditions of perfect reconstruction to design three other filters required for a 2-channel nonseparable wavelet system. Thus, \cite{gupta2011two} designs only one filter matched to the class of images. Also, the designed wavelets are matched to images, while this paper focuses on designing a wavelet that is matched to a given image.

This paper proposes a novel methodology that designs two-channel nonseparable wavelet from a given image using the convolutional neural network (CNN). The design methodology is simple and  easy to use for the design of image-matched nonseparable wavelet. Since a two-channel wavelet is designed, CNN has an architecture similar to a 2 channel filterbank. The squared error between the input image and the CNN output image is used as the loss function and is propagated back until the loss function falls considerably low or the PSNR of the reconstructed image increases beyond 70dB.

This paper is organized in four sections. Section 2 briefly reviews the basic concepts of nonseparable wavelets. Section 3 presents the proposed convolutional neural network based approach for the design of image-matched two channel non-separable wavelet. This section also presents simulation results on some images. In the end, conclusions are presented in section 4.

\textit{Notations:} We use lowercase bold letters and uppercase bold letters to represent 2-dimensional (2-D) vectors and $2 \times 2$ matrices, respectively. Thus, a discrete 2-D signal is denoted by $a[n]$ where $n=(n_1,n_2)$. The scalar variables are represented by lowercase italicized letters.

\section{Brief Background on Nonseparable Wavelets}
\label{sec:nonsep}

Consider a 2-band nonseparable wavelet system shown in Figure-1. 
\begin{figure*}[!ht]
\vspace{1em}
\label{Figure-1}
\begin{center}
\includegraphics[scale=0.55, trim =6mm 12mm 6mm 1mm]{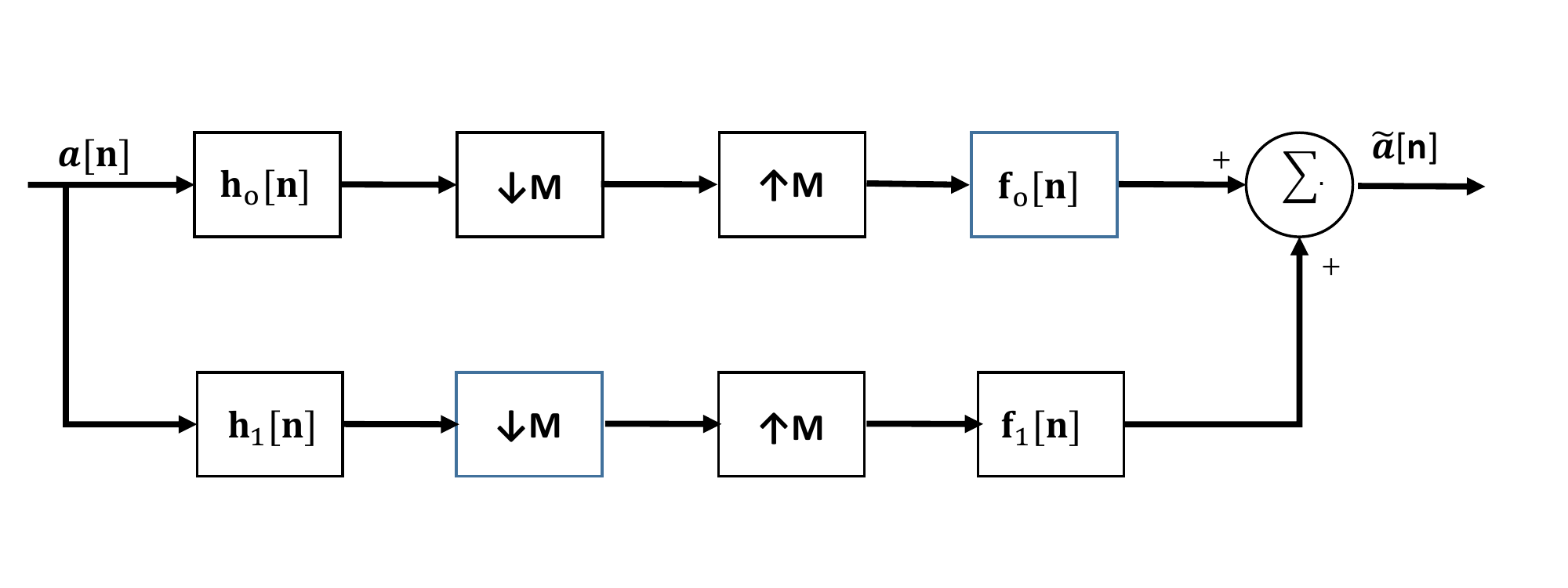}
\caption*{Figure-1: 2-Channel nonseparable wavelet system}
\end{center}
\vspace{-2em}
\end{figure*}
A two channel nonseparable wavelet system consists of four 2-dimensional filters, labeled as, $\textbf{h}_0$, $\textbf{h}_1$, $\textbf{f}_0$, and $\textbf{f}_1$, where $\textbf{h}_0$ and $\textbf{h}_1$ are analysis lowpass and highpass filters and, $\textbf{f}_0$ and $\textbf{f}_1$ are synthesis lowpass and highpass filters, respectively. In the multiresolution approximation corresponding to wavelet system, the scaling subspace is spanned by 2-D integer translates of the associated 2-D scaling function, $\Phi(\textbf{x})$, defined at that subspace \cite{rao1998wavelet}. Similar to the 1-D case, the 2-D scaling function satisfies the recursive, two-scale relationship of (1). If it exists, then the wavelet function, $\Psi(\textbf{\textbf{x}})$, satisfies (2). 
\begin{equation}
\Phi(\textbf{x})=\sum\limits_{\textbf{n}}\sqrt{|det(\textbf{M})|}\textbf{f}_0(\textbf{n})\Phi(\textbf{Mx-n})
\end{equation}
\begin{equation}
\Psi(\textbf{x})=\sum\limits_{\textbf{n}}\sqrt{|det(\textbf{M})|}\textbf{f}_1(\textbf{n})\Psi(\textbf{Mx-n})
\end{equation}
where \textbf{M} is the decimation matrix characterizing the sampling process. Likewise, the dual scaling function $\tilde{\Phi}(\textbf{\textbf{x}})$ and the dual wavelet function $\tilde{\Psi}(\textbf{\textbf{x}})$ are related to the analysis filters $\textbf{h}_0(\textbf{n})$ and $\textbf{h}_1(\textbf{n})$ via similar equations.  If $|det(\textbf{M})| =2$, then we obtain a 2-channel nonseparable wavelet system. The theory of nonseparable wavelets requires the use of lattices, where a lattice is the set of all vectors generated by \textbf{Mn}, $\textbf{n} \in Z^2$. For the 2-channel case, only quincunx lattice generates MRA and the corresponding decimation matrix generates a rotated rectangular grid. For example, matrix $\textbf{M}=[1 \ 1;1 -1]$ corresponds to the decimation matrix for the quincunx lattice.   

\section{Proposed Work on Nonseparable Wavelet Design using CNN Architecture}
\label{Section for Proposed Method}
In this section, we present the proposed design on image-matched nonseparable wavelet. We use the convolutional neural network architecture for quincunx wavelet design. 

\subsection{Proposed Design}
Consider Figure-2 that shows the CNN architecture based proposed wavelet design. The CNN network has two layers of two filters each. Each layer is trained with the gradient of loss function received from the output via back propagation. The loss function is setup as the squared error between the input applied image and the output reconstructed image as shown in Figure-2. After layer-1 output, an operation similar to pooling is applied wherein alternate image samples are replaced with zeros (Figure-2), implementing downsampling and upsampling by \textbf{M} of the corresponding wavelet structure. Layer-1 filters correspond to analysis filters and layer-2 filters correspond to synthesis filter.  

The steps for the design are as below:

\begin{enumerate}
\item The given image is applied as the input to the CNN architecture. 
\item All the filters are initialized as shown in Table-1.
\begin{figure*}[!ht]
\vspace{1em}
\label{Filters_in}
\begin{center}
\caption*{Table-1: Filter Initialization}
\includegraphics[scale=0.45, trim =6mm 6mm 6mm 1mm]{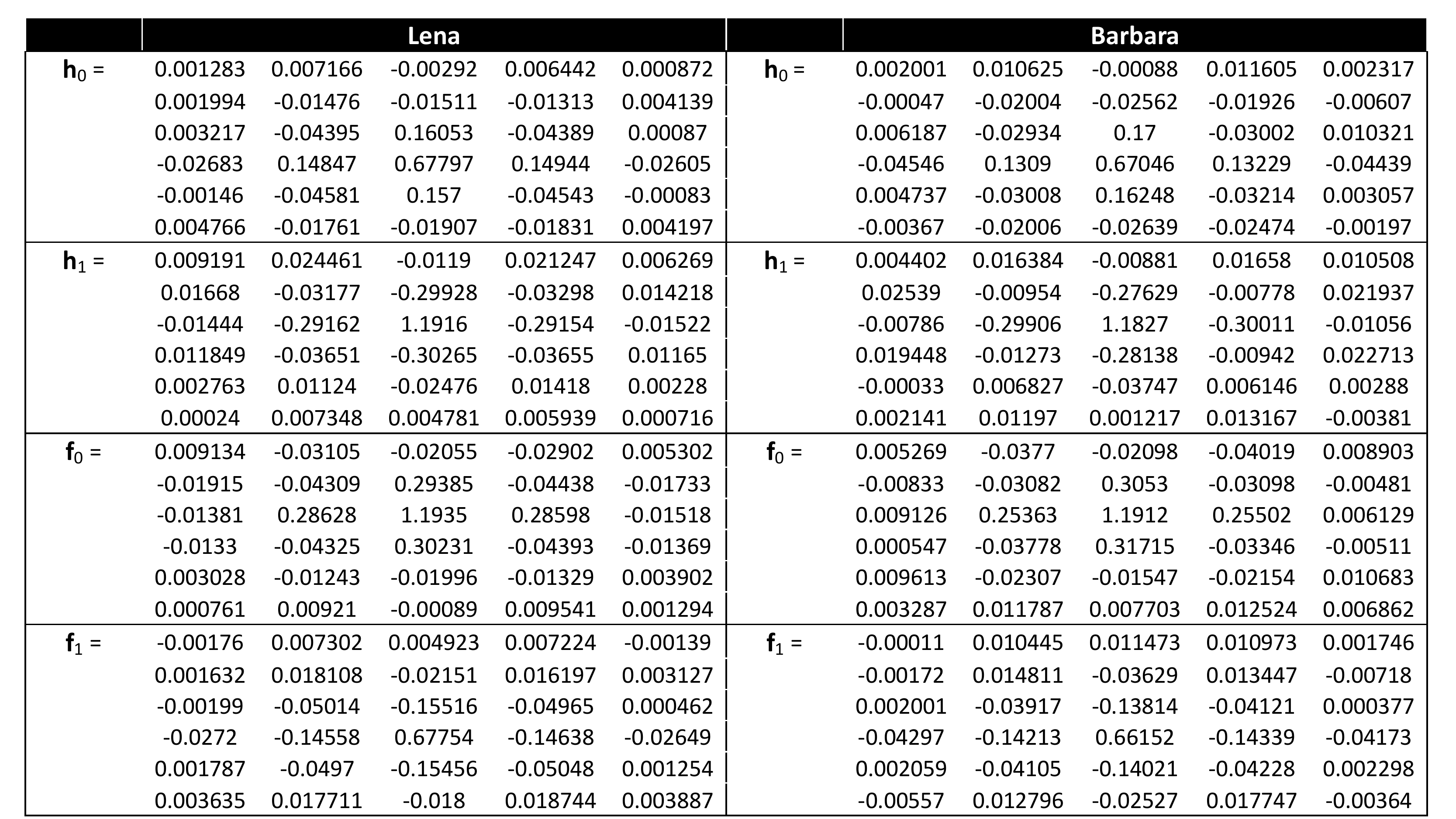}
\end{center}
\end{figure*}
A 2-channel wavelet system is implemented with lowpass and highpass filters. Hence, instead of a random initialization, filters have been initialized to small-sized lowpass and highpass filters. 

\item Loss function at the output is computed as 
\begin{equation}
L=\sum\limits_{\textbf{n}}(\textbf{a}[\textbf{n}]-\tilde{\textbf{a}}[\textbf{n}])^2
\end{equation}
where $\textbf{a}[\textbf{n}]$ denotes the input image, $\tilde{\textbf{a}}[\textbf{n}]$ denotes the reconstructed image, and the summation is done for the entire size of the input image. In this work, we have considered two $512 \times 512$ sized raw BMP images.

\item Once initialized, the CNN training is started wherein the gradient of loss function is propagated back to all the four filters and the network is trained using back-propagation (BP) algorithms leading to perfect reconstruction at the output at the end of training.  
\end{enumerate}
\begin{figure*}[!ht]
\label{method}
\begin{center}
\includegraphics[scale=0.4, trim =6mm 6mm 6mm 1mm]{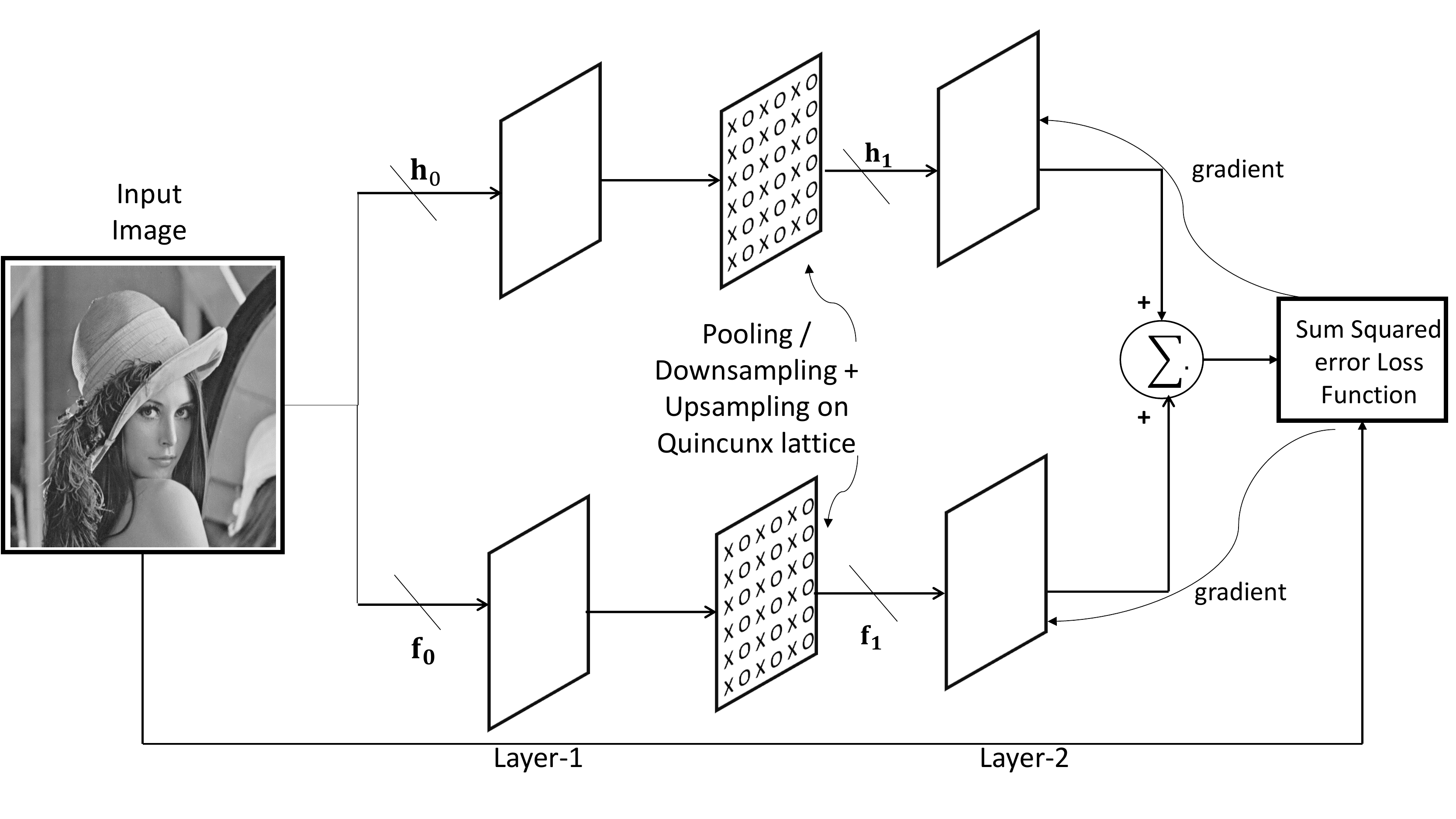}
\end{center}
\caption*{Figure 2: Proposed CNN Architecture for Nonseparable Wavelet Design over Quincunx Lattice\\ 
`x': denotes sample kept, `o' denotes sample discarded}
\end{figure*}

\subsection{Simulation Results}
Results are presented over two images: Lena and Barbara shown in Figure-3. 
\begin{figure}[!ht]
\label{images}
\begin{center}
\includegraphics[scale=0.26, trim =6mm 6mm 6mm 1mm]{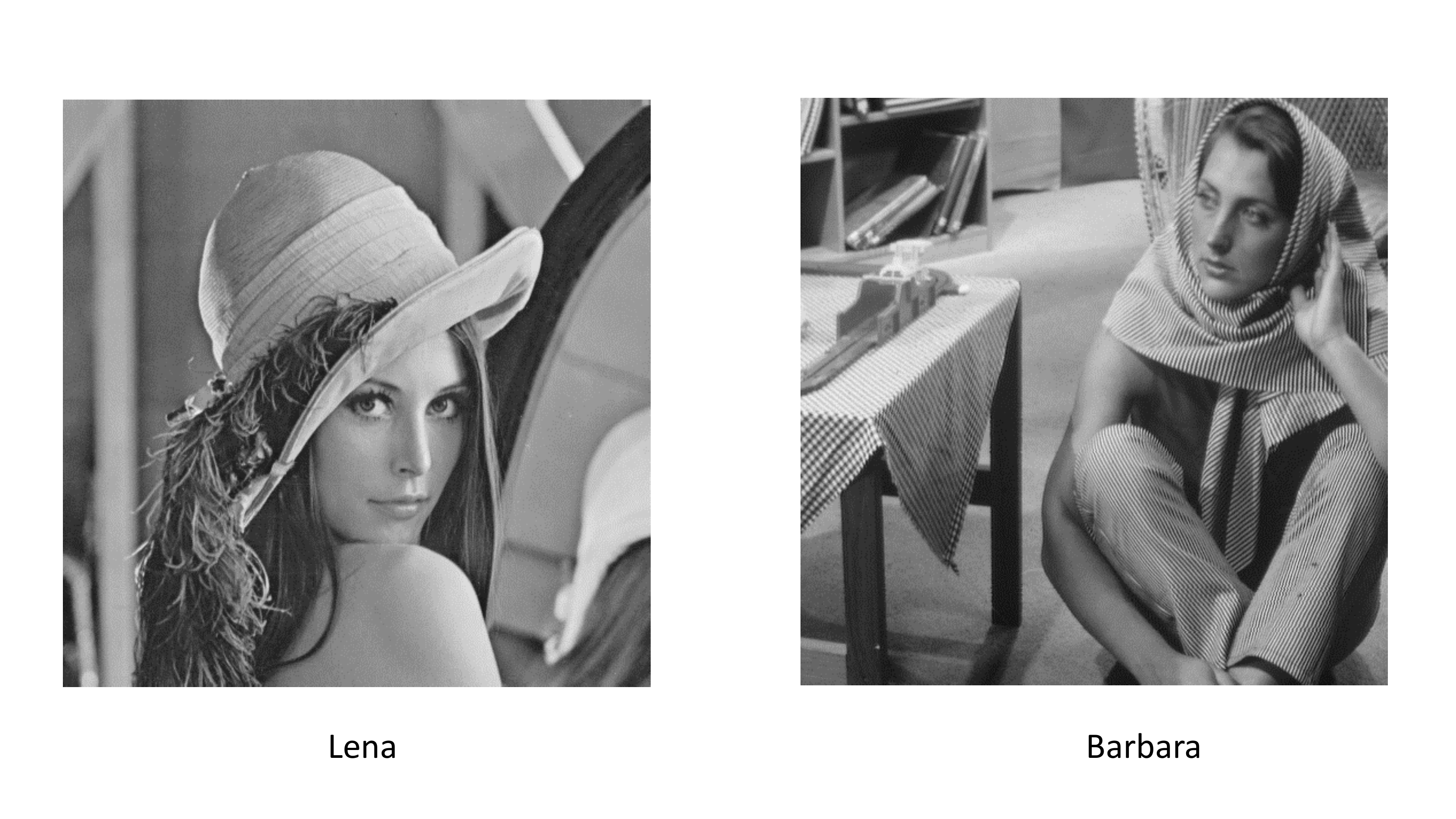}
\end{center}
\caption*{Figure-3: Images Used}
\end{figure}

The CNN network is trained using the learning rate=$2 \times 10^{-7}$, using Stochastic Gradient Descent (SGD) with Nesterov momentum with momentum value equal to 0.9. The network got trained for each image in approx. 15000 iterations. Perfect reconstruction is obtained  at the end of training, i.e., when the loss function dropped to a very low insignificant value. The designed filters are tabulated in Table-2 and the corresponding filters are shown in Figure-3.
\begin{figure*}[!ht]
\vspace{1em}
\label{Filters}
\begin{center}
\caption*{Table-2: Coefficients of image-matched filters designed using the proposed methodology}
\includegraphics[scale=0.5, trim =6mm 6mm 6mm 1mm]{Filters.pdf}
\end{center}
\end{figure*}
\begin{figure*}[!ht]
\label{Filters2}
\begin{center}
\includegraphics[scale=0.6, trim =6mm 6mm 6mm 1mm]{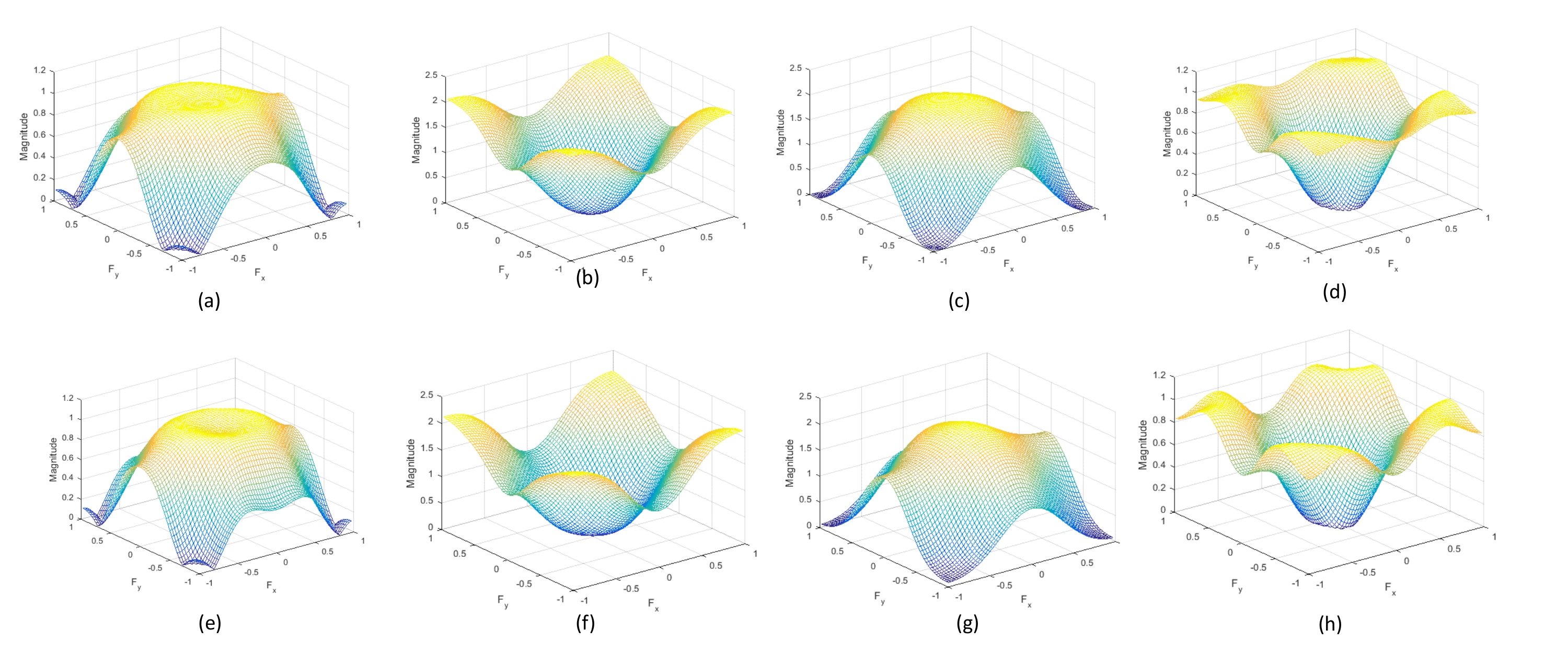}
\end{center}
\caption*{Figure-4: Image-matched filters designed using the proposed methodology; Lena image: (a) $\textbf{h}_0$, (b) $\textbf{h}_1$, (c) $\textbf{f}_0$, and (d) $\textbf{f}_1$; Barbara Image: (e) $\textbf{h}_0$, (f) $\textbf{h}_1$, (g) $\textbf{f}_0$, and (h) $\textbf{f}_1$}
\end{figure*}
\section{Conclusion}
This paper proposes a novel methodology for designing a two-channel nonseparable wavelet over quincunx lattice from a given image. Since this wavelet is designed from the given image itself, it may be able to capture image characteristics/features better than the existing wavelets. Hence, this design methodology can find potential use in image processing applications. In the future, this design will be extended to 3-dimensional wavelet design and will be explored in applications of classification and segmentation.   
\bibliographystyle{IEEEtran}
\bibliography{refs}

\end{document}